\documentclass{opt2020} % Include author names
\title[Stochastic Damped L-BFGS with Controlled Norm of the Hessian Approximation]{Stochastic Damped L-BFGS with Controlled Norm of the Hessian Approximation}

\optauthor{%
    \Name{Sanae Lotfi} \Email{sanae.lotfi@polymtl.ca}\\
    \Name{Tiphaine Bonniot de Ruisselet} \Email{tiphaine.bonniot@gmail.com}\\
    \Name{Dominique Orban} \Email{dominique.orban@polymtl.ca}\\
    \Name{Andrea Lodi} \Email{andrea.lodi@polymtl.ca}\\
    \addr Polytechnique Montréal, Canada
}

\usepackage{gitinfo}

\newcommand{\cahiernumber}{52}  % Insert your Cahier du GERAD number.
\makeatletter
\def\cahierline{\gdef\@cahierline}
\cahierline{\textcolor{gray}{Cahier du GERAD G-\number\year-\cahiernumber}}
\def\gitline{\gdef\@gitline}
\gitline{\textcolor{gray}{Commit \gitAbbrevHash\ by \gitAuthorName\ on \gitAuthorIsoDate}}
\def\ps@firstpage{%\ps@plain
\def\@oddfoot{\normalfont\scriptsize \hfil\thepage\hfil\@gitline}%
  \let\@evenfoot\@oddfoot
  \def\@oddhead{\hfil\normalfont\small\@cahierline}%
  \let\@evenhead\@oddhead % in case an article starts on a left-hand page
}

\def\ps@myheadings{%
    \def\@oddfoot{\normalfont\scriptsize\ttfamily\@cahierline\hfil\@gitline}
    \def\@evenfoot{\normalfont\scriptsize\ttfamily\@gitline\hfil\@cahierline}
    \let\@mkboth\@gobbletwo
    }
\makeatother

\usepackage[font=small]{caption}

\usepackage{amsfonts}
\newcommand{\E}{\mathbb{E}}
\newcommand{\R}{\mathbb{R}}

\newcommand{\BigO}{\mathcal{O}}

\usepackage{amsmath}

\usepackage{mathtools}
\usepackage{subdepth}

\usepackage{pifont}% http://ctan.org/pkg/pifont
\usepackage[textwidth=2.5cm,textsize=tiny]{todonotes}

\usepackage{algorithm}
\usepackage[noend]{algorithmic}

\newtheorem{assumption}{Assumption}

\usepackage{enumitem,amssymb}
\newlist{todolist}{itemize}{2}
\setlist[todolist]{label=$\square$}

\usepackage[nameinlink]{cleveref}
\crefname{assumption}{assumption}{assumptions}

\begin{document}

\maketitle

\thispagestyle{firstpage}
\pagestyle{myheadings}

\begin{abstract}%
We propose a new stochastic variance-reduced damped L-BFGS algorithm, where we leverage estimates of bounds on the largest and smallest eigenvalues of the Hessian approximation to balance its quality and conditioning.
Our algorithm, VARCHEN, draws from previous work that proposed a novel stochastic damped L-BFGS algorithm called SdLBFGS.
We establish almost sure convergence to a stationary point and a complexity bound.
We empirically demonstrate that VARCHEN is more robust than SdLBFGS-VR and SVRG on a modified DavidNet problem---a highly nonconvex and ill-conditioned problem that arises in the context of deep learning, and their performance is comparable on  a logistic regression problem and a nonconvex support-vector machine problem.
% A variance-reduced version of SdLBFGS-CHN improves the quality of the curvature approximation and accelerates convergence.
\end{abstract}

%\begin{keywords}%
%  List of keywords%
%\end{keywords}

\section{Introduction and Related Work}

We consider unconstrained stochastic minimization problems of $f: \R^n \to \R$ where
% \begin{equation}
% \label{eq:optim-prob-4}
% \min_{x \in \R^n} f(x) := \E_\xi[F(x,\xi)],
% \end{equation}
\begin{equation}
\label{eq:optim-prob}
    f(x) = \E_\xi[F(x,\xi)]
    \ \text{(online)} \quad \text{or} \quad
    f(x) = \frac{1}{N} \sum_{i=1}^{N} f_i(x)
    \ \text{(finite sum)},
% \min_{x \in \R^n} f(x) :=
% \begin{cases}
%      \E_\xi[F(x,\xi)],
%     & \text{\textit{online}}\\
%   \frac{1}{N} \sum_{i=1}^{N} f_i(x),
%     & \text{\textit{finite-sum}} \\
% \end{cases}
\end{equation}
where $\xi \in \R^d$ denotes a random variable, \(F : \R^n \times \R^d \to \R\) is continuously differentiable and possibly nonconvex, $f_i$ is the loss corresponding to the $i$-th element of our dataset and $N$ is the size of the dataset.
The algorithm developed below applies to both online and finite-sum problems.

% also called \textit{empirical risk minimization} problems.
% \todo{why not call $f$ empirical risk in \eqref{eq:optim-prob}?}
% We introduce additional assumptions about \(F\) as needed. In the context of machine learning, a special case of the optimization problem~\eqref{eq:optim-prob-4} is \textit{empirical risk minimization}, which consists of minimizing a different form of $f$,

% \begin{equation}
% \label{eq:stoch-optim-prob-4}
% \min_{x \in \R^n} f(x) :=
% \left \{
%     \begin{array}{l l}
%      \E_\xi[F(x,\xi)],
%     & \text{if $k=0$}\\
%   \frac{1}{N} \sum_{i=1}^{N} f_i(x),
%     & \text{otherwise} \\
%     \end{array}
% \right.
% \end{equation}

% where $f_i$ represents the loss function that corresponds to the $i$-th element of our dataset and $N$ is the size of the dataset. Our algorithm applies to both online and finite-sum problems.

Stochastic Gradient Descent (SGD)~\citep{robbins1951stochastic, bottou2010large} and its variants~\citep{polyak1964some, nesterov1983method, duchi2011adaptive, tieleman2012rmsprop, Adam}, including variance-reduced algorithms~\citep{svrg,nguyen2017sarah,fang2018spider, wang2019spiderboost}, are widely used to solve~\eqref{eq:optim-prob} in machine learning.
However, they might not be well-suited for highly nonconvex and ill-conditioned problems~\citep{bottou2018optimization}, which are more effectively treated using (approximate) second-order information.
Second-order algorithms are well studied in the deterministic case~\citep{dennis1974characterization, dembo1982inexact, dennis1996numerical, amari1998natural} but there are many areas to explore in the stochastic context that go beyond existing works~\citep{schraudolph2007stochastic, bordes2009sgd, byrd2016stochastic, moritz2016linearly,gower2016stochastic}.
Among these areas, the use of damping in L-BFGS is an interesting research direction to be leveraged in the stochastic case.
\citet{wang2017stochastic} proposed a stochastic damped L-BFGS (SdLBFGS) algorithm and proved almost sure convergence to a stationary point.
However, damping does not prevent the inverse Hessian approximation \(H_k\) from being ill-conditioned \citep{chen2019stochastic}.
The convergence of SdLBFGS may be heavily affected if the Hessian approximation becomes nearly singular during the iterations.
% SdLBFGS does not guarantee that it will not be the case.
In order to remedy this issue, \citet{chen2019stochastic} proposed to combine SdLBFGS with regularized BFGS \cite{mokhtari2014res}.
Our approach differs.

\section*{Our contributions:}

\begin{itemize}
    \setlength{\itemsep}{0pt}
    \item Less restrictive assumptions: we force \(H_k\) to be uniformly bounded and positive definite by requiring the stochastic gradient to be Lipschitz continuous, which is a less restrictive assumption than those of \citet{wang2017stochastic}, who require the stochastic function to be twice differentiable with respect to the parameter vector, and its Hessian to be bounded for all parameter and random sampling vectors.
    \item A new damped L-BFGS: we propose a new version of stochastic damped L-BFGS that maintains estimates of the smallest and largest eigenvalues of \(H_k\).
    % to change the update of the inverse Hessian approximation.
    A solution is proposed when ill-conditioning is detected that preserves almost sure convergence to a stationary point.
    \item Choice of the initial inverse Hessian approximation: we propose a new formula for the initial inverse Hessian approximation to make the algorithm more robust to ill-conditioning.
    % \todo{``more stable'' is very vague}
\end{itemize}

\paragraph{Notation}
For a symmetric matrix \(A\), we use \(A \succ 0\) to indicate that \(A\) is positive definite, and \(\lambda_{\min}(A)\) and \(\lambda_{\max}(A)\) to denote its smallest and largest eigenvalue, respectively.
If \(B\) is also symmetric, $B \preceq A$ means that $A - B$ is positive semidefinite.
The identity matrix of appropriate size is denoted \(I\). Finally, $\E_\xi[.]$ is the expectation over random variable $\xi$.

% \subsection*{Notation}

\section{Formulation of our method}

We assume that %we do not have access to full gradients, but that
at iteration \(k\), we can obtain a stochastic approximation
\begin{equation}
\label{eq:mini-batch-grad}
 g(x_k, \xi_k) = \frac{1}{m_k} \sum_{i=1}^{m_k} \nabla f_{\xi_{k,i}}(x_k)
\end{equation}
of \(\nabla f(x_k)\), where $\xi_{k}$ denotes the subset of samples taken from a given set of realizations of $\xi$.
%
% where $\xi_{k,i}$ is to the $i$-th sample of $\xi_{k}$ in iteration $k$ and $m_k$ is the batch size used in iteration $k$.
% Note that we do not assume here that $g(x_k, \xi_{k,i}) = \nabla f_{\xi_{k,i}}(x_k)$.
%
The Hessian approximation constructed at iteration $k$ and its inverse are denoted by $B_k$ and $H_k$, respectively, such that $H_k = B_k^{-1}$ and $B_k \succ 0$.
Iterates are updated according to
\begin{equation}
\label{eq:update}
  x_{k+1} = x_k + \alpha_k d_k, \quad \text{where} \quad d_k = -H_k g(x_k, \xi_k) \quad \text{and} \quad  \alpha_k > 0 \, \text{is the step size.}
\end{equation}

The stochastic BFGS method~\citep{schraudolph2007stochastic} computes an updated approximation \(H_{k+1}\) according to
\begin{equation}
  \label{eq:BFGS-update_4}
  H_{k+1} = V_k H_k V_k^\top + \rho_k s_k s_k^\top,
  \quad \text{where } \quad
  V_k = I - \rho_k s_k y_k^\top
  \quad \text{and} \quad
  \rho_k = 1 / s_k^\top  y_k,
\end{equation}
which ensures that the secant equation $B_{k+1}s_k = y_k$ is satisfied, where
\begin{equation}
\label{eq:define-y-s}
  s_k = x_{k+1}-x_k, \quad \text{and} \quad
  y_k = g(x_{k+1}, \xi_k) - g(x_k, \xi_k).
\end{equation}
If \(H_k \succ 0\) and the curvature condition $s_k^\top y_k > 0$ holds, then $H_{k+1} \succ 0$~\citep[see, for instance,][]{fletcher1970new}.
% The initial approximation $H_0$ is chosen to be positive definite. Recursively, $H_{k+1}$ is positive definite .

Because storing \(H_k\) and performing matrix-vector products is costly for large-scale problems, we use the limited-memory version of BFGS (L-BFGS)~\citep{nocedal1980updating, liu1989limited}, in which \(H_k\) only depends on the most recent $p$ iterations and an initial $H_k^0 \succ 0$.
The parameter $p$ is the \emph{memory} of L-BFGS.
The inverse Hessian update can be written as
% \todo{reference Nocedal's paper}
\begin{equation}
\label{eq:LBFGS-update_4}
\begin{aligned}
  H_k = &\ (V_{k-1}^\top  \dots V_{k-p}^\top ) H_k^0 (V_{k-p} \dots V_{k-1}) + \\
        & \rho_{k-p}(V_{k-1}^\top  \dots V_{k-p+1}^\top ) s_{k-p} s_{k-p}^\top  (V_{k-p+1} \dots V_{k-1}) + \dots + % \\
        % &+\rho_{k-p+1}(V_{k-1}^\top  \dots V_{k-p+2}^\top ) s_{k-p+1} s_{k-p+1}^\top  (V_{k-p+2} \dots V_{k-1})  \\
        % & +
        \rho_{k-1} s_{k-1}s_{k-1}^\top. % , \quad \text{where} \quad V_k = (I - \rho_k s_k y_k^\top ).
\end{aligned}
\end{equation}
When $\alpha_k$ in~\eqref{eq:update} is not computed using a Wolfe line search \citep{wolfe1969convergence, wolfe1971convergence}, there is no guarantee that the curvature condition holds.
A common strategy is to simply skip the update.
By contrast, \citet{powell1978algorithms} proposed \textit{damping}, which consists in updating \(H_k\) using a modified \(y_k\), denoted by $\hat{y}_k$, to benefit from information discovered at iteration \(k\) while ensuring sufficient positive definiteness.
% Our purpose is not only to prove that the inverse Hessian approximation, $H_k$, stays bounded all along the optimization process, but also to obtain bounds that we are able to compute, or at least compute their approximations.
% , since the latter would result in an expression of the bounds over $H_{k+1}$ that depends on the maximum and minimum eigenvalues of $H_{k}$, which we cannot afford to compute.
We use
\begin{equation}
\label{eq:new-ytilde}
  \hat{y}_k := \theta_k y_k + (1-\theta_k) B_{k+1}^0 s_k,
\end{equation}
which is inspired by~\citep{wang2017stochastic}, and differs from the original proposal of \citet{powell1978algorithms}, where
\begin{equation}
\label{eq:new_theta}
\theta_k =
1 \ \text{ if }\ s_k^\top y_k \geq \eta s_k^\top  B_{k+1}^0 s_k, \ \text{ and }\ (1-\eta)\frac{s_k^\top  B_{k+1}^0 s_k}{s_k^\top  B_{k+1}^0 s_k - s_k^\top  y_k} \text{ otherwise,}
% \begin{cases}
%   1
%   & \text{if} \quad s_k^\top y_k \geq \eta s_k^\top  B_{k+1}^0 s_k \\
%   (1-\eta)\frac{s_k^\top  B_{k+1}^0 s_k}{s_k^\top  B_{k+1}^0 s_k - s_k^\top  y_k}
%   & \text{if} \quad s_k^\top y_k < \eta s_k^\top  B_{k+1}^0 s_k,
% \end{cases}
\end{equation}
with $\eta \in (0,1)$ and \(B_{k+1}^0 := (H_{k+1}^0)^{-1}\).
The choice~\eqref{eq:new-ytilde} ensures that the curvature condition
\begin{equation}
\label{eq:new-sTy-bounded}
    s_k^\top  \hat{y}_k \geq \eta s_k^\top  B_{k+1}^0 s_k \geq \eta \lambda_{\min}(B_{k+1}^0) \|s_k\|^2 > 0 ,
    % = \frac{\eta}{\lambda_{\max}(H_{k+1}^0)} \|s_k\|^2,
\end{equation}
is always satisfied since $H_{k+1}^0 \succ 0$.
We obtain the damped L-BFGS update,
% \begin{equation}
%   \label{eq:damped-BFGS-update}
%   H_{k+1} = \hat{V}_k H_k \hat{V}_k^\top + \hat{\rho}_k s_k s_k^\top,
%   \quad \text{where } \quad
%   \hat{V}_k = I - \hat{\rho}_k s_k^\top \hat{y}_k
%   \quad \text{and} \quad
%   \hat{\rho}_k = 1 / s_k^\top  \hat{y}_k.
% \end{equation}
% The L-BFGS update with damping
which is~\eqref{eq:LBFGS-update_4} with each \(V_i\) and \(\rho_i\) replaced with \(\hat{V}_i = I - \hat{\rho}_i s_i^\top \hat{y}_i\) and \(\hat{\rho}_i = 1 / s_i^\top  \hat{y}_i\).

\section{A new stochastic damped L-BFGS with controlled Hessian norm}

Our working assumption is
\begin{assumption}
  \label{asm:main}
  There is \(\kappa_{\text{low}} \in \R\) such that \(f(x) \geq \kappa_{\text{low}}\) for all \(x \in \R^n\),
  \(f\) is \(\mathcal{C}^1\) over \(\R^n\), and
  there is \(L > 0\) such that for all \(x\), \(y \in \R^n\), \(\|\nabla f(x) - \nabla f(y)\| \leq L \, \|x - y\|\).
\end{assumption}

% \begin{assumption}
%   \label{asm:f-bounded_4}
%   There exists \(\kappa_{\text{low}} \in \R\) such that \(f(x) \geq \kappa_{\text{low}}\) for all \(x \in \R^n\).
% \end{assumption}

% \begin{assumption}
%   \label{asm:f-C1_4}
%   The function \(f\) is continuously differentiable over \(\R^n\).
% \end{assumption}

% \begin{assumption}
%   \label{asm:g-lipschitz_4}
%   There exists \(L > 0\) such that for all \(x\), \(y \in \R^n\), \(\|\nabla f(x) - \nabla f(y)\| \leq L \, \|x - y\|\).
% \end{assumption}

% In \Cref{sec:bounds}, we provide bounds on the smallest and largest eigenvalues of \(H_{k+1}\) as functions of bounds on those of \(H_k\).
% In \Cref{sec:algorithm_4}, we present our algorithm: Variance-Reduced Stochastic Damped L-BFGS with Controlled Hessian Norm (VR-SdLBFGS-CHN).
% \todo{do we really want to go with that name?}

% Finally, in Section~\ref{sec:VR-algorithm}, we propose a variance reduced version of SdLBFGS-CHN.

% \subsection{Bounded inverse Hessian approximation}\label{sec:bounds}

We begin by deriving bounds on the smallest and largest eigenvalues of \(H_{k+1}\) as functions of bounds on those of \(H_k\).
Proofs can be found in \Cref{app:proofs}.

% \begin{restatable}{lemma}{firstlemma}
\begin{lemma}
  \label[lemma]{lem:eigenvalues-update}
  Let $s$ and $y \in \mathbb{R}^n$ such that $s^\top y \geq \gamma \|s\|^2$ with $\gamma > 0$, and such that $\|y\|\leq L_y \|s\|$, with $L_y > 0$.
  Let $A =\mu V V^\top  + \rho s s^\top $, where $\rho = 1 / s^\top y$, $\mu > 0$, and $V = I - \rho s y^\top$.
%   \todo{$S$ should be $V$}
  Then,
  \[
    0 <
    \min \left( \frac{1}{L_y}, \frac{\mu}{1 + \frac{\mu}{\gamma} L_y^2} \right) \leq
    \lambda_{\min}(A) \leq
    \lambda_{\max}(A) \leq
    \frac{1}{\gamma} + \max \left( 0, \frac{\mu}{\gamma^2}L_y^2 - \frac{\mu}{1 + \frac{\mu}{\gamma}L_y^2} \right).
  \]
%   \begin{align}
%     \lambda_{\min}(A) & \geq \min \left( \frac{1}{L_y}, \frac{\mu}{1 + \frac{\mu}{\gamma} L_y^2} \right) > 0,
%     \label{eq:lambda-min}
%     \intertext{and}
%     \lambda_{\max}(A) & \leq \frac{1}{\gamma} + \max \left( 0, \frac{\mu}{\gamma^2}L_y^2 - \frac{\mu}{1 + \frac{\mu}{\gamma}L_y^2} \right).
%     \label{eq:lambda-max}
%   \end{align}
\end{lemma}

To use \Cref{lem:eigenvalues-update} to obtain bounds on the eigenvalues of \(H_{k+1}\), we make the following assumption:
\begin{assumption}
\label{asm:stochastic-g-lipschitz}
%   The approximation $g(x, \xi)$ is Lipschitz continuous, i.e.,
  There is \(L_g > 0\) such that for all \(x\), \(y \in \R^n\), \(\|g(x, \xi) - g(y, \xi)\| \leq L_g \, \|x - y\|\).
\end{assumption}

\Cref{asm:stochastic-g-lipschitz} is required to prove convergence and convergence rates for most recent stochastic quasi-Newton methods~\citep{yousefian2017smoothing}.
It is less restrictive than requiring $f(x,\xi)$ to be twice differentiable with respect to $x$, and the Hessian $\nabla_{xx}^{2}f(x,\xi)$ to be bounded for any $x$, $\xi$, as in~\citep{wang2017stochastic}.

The next theorem shows that the eigenvalues of $H_{k+1}$ are bounded and bounded away from zero.
% \todo{What are the assumptions of this thm?}

\begin{theorem}
  \label{thm:new-L-BFGS-update}
Let \Cref{asm:main,asm:stochastic-g-lipschitz} hold. Let $H_{k+1}^0 \succ 0$ and $p > 0$.
If $H_{k+1}$ is obtained by applying $p$ times the damped BFGS update formula with inexact gradient to $H_{k+1}^0$, there exist easily computable constants $\lambda_{k+1}$ and $\Lambda_{k+1}$ that depend on $L_g$ and $H_{k+1}^0$ such that \(0 < \lambda_{k+1} \leq \lambda_{\min}(H_{k+1}) \leq \lambda_{\max}(H_{k+1})\leq \Lambda_{k+1}\).

\end{theorem}

% Now that we obtained the upper~(\ref{eq:lambdamax2}) and lower~(\ref{eq:lambdamin}) bounds on the maximum and minimum eigenvalues of $H_{k+1}$, respectively, we use them to define our algorithm in the next section.
The precise form of \(\lambda_{k+1}\) and \(\Lambda_{k+1}\) is given in \Cref{app:proofs}.
% We use the bounds of \Cref{thm:new-L-BFGS-update} to define our algorithm in the next section.

% \subsection{Description of the algorithm: SdLBFGS-CHN }\label{sec:algorithm_4}

% First, we discuss the choice of $H_{k+1}^0$.
A common choice for $H_{k+1}^0$ is
% \todo{Cite papers instead of books}
\(
    H_{k+1}^0 = \gamma_{k+1}^{-1} I %, \quad \text{where} \quad \gamma_{k+1} = y_{k}^\top y_k / s_k^\top y_k,
\)
where \(\gamma_{k+1} = y_{k}^\top y_k / s_k^\top y_k\),
is the \emph{scaling parameter}. This choice ensures that the search direction is well scaled, which promotes large steps.
To keep $H_{k+1}^0$ from becoming nearly singular or non positive definite, % (since there is no guarantee in the nonconvex case that $s_k^\top y_k > 0$),
we define
\begin{equation}
\label{eq:init_HK}
    H_{k+1}^0 = \left( \max(\underline{\gamma}_{k+1}, \min(\gamma_{k+1},\overline{\gamma}_{k+1})) \right) I, %\quad \text{where} \quad \gamma_{k+1} = y_{k}^\top y_k / s_k^\top y_k,
\end{equation}
where $0 < \underline{\gamma}_{k+1} < \overline{\gamma}_{k+1}$ can be constants or iteration dependent.

% We use the mini-batch gradient approximation

% \begin{equation}
% \label{eq:mini-batch-grad}
%  g(x_k, \xi_k) = \frac{1}{m_k} \sum_{i=1}^{m_k} \nabla f_{\xi_{k,i}}(x_k).
% \end{equation}

The Hessian-gradient product used to compute the search direction $d_k = - H_k g(x_k,\xi_k)$ can be obtained cheaply by exploiting a recursive algorithm \citep{nocedal1980updating}, as described in \Cref{alg:two-loop-recursion} in \Cref{app:algorithms}.

Motivated by the success of recent methods combining variance reduction with stochastic L-BFGS \citep{gower2016stochastic, moritz2016linearly, wang2017stochastic}, we apply
% \todo{This sentence and next seem to be malformed. \textcolor{blue}{AL: I think they can be fixed by adding ``we'' twice, which I did but, Sanae, please check.}}
an SVRG-like type of variance reduction~\citep{svrg} to the update.
Not only would this accelerate the convergence, since we can choose a constant step size, but it also improves the quality of the curvature approximation.

We summarize our complete algorithm,  VAriance-Reduced stochastic damped L-BFGS with Controlled HEssian Norm (VARCHEN), as \Cref{alg:VR-SdLBFGS-CHN}.
% \todo{In the algorithm, what is $H_0$?}
% The vanilla version of this algorithm is given by \Cref{alg:lbfgs-inexact} in \Cref{app:algorithms}.

\begin{algorithm}[htbp]
  \caption{Variance-Reduced Stochastic Damped L-BFGS with Controlled Hessian Norm}
  \label{alg:VR-SdLBFGS-CHN}
  \begin{algorithmic}[1]
    % \REQUIRE \(x_0 \in \R^n\)

    \STATE Choose \(x_0 \in \R^n\), step size sequence \(\{\alpha_k > 0\}_{k\geq 0}\), batch size sequence \(\{m_k > 0\}_{k\geq 0}\), eigenvalue limits $\lambda_{\max} > \lambda_{\min} > 0$, memory parameter $p$, total number of epochs $N_{\text{epochs}}$, and sequences \(\{\underline{\gamma}_{k} > 0\}_{k\geq 0}\) and \(\{\overline{\gamma}_{k+1} > 0\}_{k\geq 0}\), such that
    $0 < \lambda_{\min} < \underline{\gamma}_{k} < \overline{\gamma}_{k} < \lambda_{\max}$, for every $k \geq 0$. Set \(k = 0\) and $H_0 = I$.
    % \textcolor{red}{Can $\underline{\gamma}_{k+1} \to 0$?}

    \FOR{$t=1,\ldots,N_{\text{epochs}}$}

    \STATE Define $x_k^t = x_k$ and compute the full gradient $\nabla f(x_k^t)$.
    Set \(M = 0\).
    \label{step3:alg:VR-SdLBFGS-CHN}

    \WHILE{$M < N$}

    \STATE Sample batch $\xi_k$ of size $m_k \leq N - M$ and compute \(g(x_k,\xi_k)\) and \(g(x_k^t,\xi_k)\).
    \label{step4:alg:VR-SdLBFGS-CHN}

    \STATE Define \(\Tilde{g}(x_k,\xi_k) = g(x_k,\xi_k) - g(x_k^t,\xi_k) + \nabla f(x_k^t)\).
    \label{step5:alg:VR-SdLBFGS-CHN}

    % \STATE If $k=0$,  define $d_k = - \Tilde{g}(x_k,\xi_k)$, otherwise compute $d_k = - H_k \Tilde{g}(x_k,\xi_k)$ using Algorithm~\ref{alg:two-loop-recursion}. Then define \(x_{k+1} = x_k + \alpha_k d_k\).

    \STATE Estimate $\Lambda_k$ and $\lambda_k$ in \Cref{thm:new-L-BFGS-update}. % using~\eqref{eq:lambdamax2} and~\eqref{eq:lambdamin}.
    If $\Lambda_k > \lambda_{\max}$ or $\lambda_k < \lambda_{\min}$, delete  $s_i$, $y_i$ and $\hat{y}_i$ for $i = k-p+1,\ldots,k-2$.
    \label{step9:alg:VR-SdLBFGS-CHN}

    \STATE Compute $d_k = - H_k \Tilde{g}(x_k,\xi_k)$.
    %     \begin{equation*}
    % \label{eq:theta}
    % d_k =
    % \left \{
    % \begin{array}{l l}
    % - \Tilde{g}(x_k,\xi_k),
    % & \text{if $k=0$}\\
    % - H_k \Tilde{g}(x_k,\xi_k),
    % & \text{otherwise} \\
    % \end{array}
    % \right.
    % \end{equation*}

    \STATE Define \(x_{k+1} = x_k + \alpha_k d_k\), and compute $s_k$, $y_k$ as in~\eqref{eq:define-y-s}, and $\hat{y}_k$ as in~\eqref{eq:new-ytilde}.

    % \STATE Compute $s_k$, $y_k$ as in~\eqref{eq:define-y-s}, and $\hat{y}_k$ as in~\eqref{eq:new-ytilde} using $\Tilde{g}(x_k,\xi_k)$.

    % \STATE Estimate $\Lambda_k$ and $\lambda_k$ in~\eqref{eq:eigenvalues-bounds} recursively using~\eqref{eq:lambdamax2} and~\eqref{eq:lambdamax}. If $\Lambda_k > \lambda_{\max}$ or $\lambda_k < \lambda_{\min}$, use strategy~\ref{alg:hessian-strategy1} or~\ref{alg:hessian-strategy2}.

    \STATE Increment \(k\) by one and update \(M \leftarrow M + m_k\).
    % go to step~\ref{step4:alg:VR-SdLBFGS-CHN}.

    % \STATE If the epoch is finished, increment $t$ by one, \(k\) by one and go to step~\ref{step3:alg:VR-SdLBFGS-CHN}. Otherwise, increment \(k\) by one and go to step~\ref{step4:alg:VR-SdLBFGS-CHN}.

    \ENDWHILE

    \ENDFOR

    % \IF{the epoch is completed and $t=N_{\text{epochs}}-1$}
    % \STATE Terminate with the approximate solution \(x_{k+1}\). \\

    % \ELSIF{the epoch is completed and $t<N_{\text{epochs}}-1$}
    % \STATE Increment \(t\) by one, shuffle the dataset, increment \(k\) by one and go to step~\ref{step3:alg:VR-SdLBFGS-CHN}.
    % \ELSE
    % \STATE Increment \(k\) by one and go to step~\ref{step4:alg:VR-SdLBFGS-CHN}.
    % \ENDIF
  \end{algorithmic}
\end{algorithm}

In step~\ref{step9:alg:VR-SdLBFGS-CHN} of \Cref{alg:VR-SdLBFGS-CHN}, we compute an estimate of the upper and lower bounds on $\lambda_{\max}(H_{k})$ and $\lambda_{\min}(H_{k})$, respectively. The only unknown quantity in the expressions of $\Lambda_k$ and $\lambda_k$ in \Cref{thm:new-L-BFGS-update} is $L_g$, which we estimate as $L_g \approx L_{g,k} := \|y_k\| / \|s_k\|$.
When the estimates are not within the limits $[\lambda_{\min}, \, \lambda_{\max}]$,  % that we set as hyperparameters.
we delete $s_i$, $y_i$ and $\hat{y}_i$, $i \in \{k-p+1,\ldots,k-2\}$ from storage, such that $H_{k} g(x_{k},\xi_{k})$ is computed using the most recent pair $(s_{k-1}, \hat{y}_{k-1})$ only and $d_{k} = - H_{k} g(x_{k},\xi_{k})$.
% \todo{We need $\lambda_{\min} I \preceq H_{k+1}^0 \preceq \lambda_{\max} I$}
% Note that the notation $\lambda_k$ and $\Lambda_k$ do not refer to the $k$-th eigenvalue and that the index $k$ refers to the iteration number only.
Finally, a full gradient is computed once in every epoch in step~\ref{step3:alg:VR-SdLBFGS-CHN}.
The term $g(x_k^t,\xi_k) - \nabla f(x_k^t)$ can be seen as the bias in the gradient estimation $ g(x_k,\xi_k)$, and it is used here to correct the gradient approximation in step~\ref{step5:alg:VR-SdLBFGS-CHN}.

\section{Convergence and Complexity Analysis}
\label{sec:convergence_4}

We show that \Cref{alg:VR-SdLBFGS-CHN} satisfies the assumptions of the convergence analysis and iteration complexity of~\citet{wang2017stochastic} for stochastic quasi-Newton methods.
% We consider \Cref{asm:f-bounded_4,asm:f-C1_4,asm:g-lipschitz_4}.
We make an additional assumption used by~\citet{wang2017stochastic} to establish global convergence.
% \todo{Define $\E_{\xi_k}$}

\begin{assumption}
  \label{asm:bounded-exp}
  For all $k$, $\xi_k$ is independent of $\{x_1, \ldots, x_k\}$, \(\E_{\xi_k}\left[ g(x_k, \xi_k) \right] = \nabla f(x_k)\), and there exists $\sigma > 0$ such that \(\E_{\xi_k}[ \| g(x_k, \xi_k) - \nabla f(x_k) \|^2 ] \leq \sigma^2\).
%   \begin{align}
%   \label{eq:unbiased-grad-app}
%       \E_{\xi_k}\left[ g(x_k, \xi_k) \right] & = \nabla f(x_k),  \\
%     %   \intertext{and}
%       \E_{\xi_k}[ \| g(x_k, \xi_k) - \nabla f(x_k) \|^2 ] & \leq \sigma^2.
%   \end{align}
\end{assumption}

% \begin{assumption}
%   \label{asm:bounded-hessian}
%   There exist $\Lambda \geq \lambda > 0$ such that $\lambda I \preceq H_k \preceq \Lambda I$ for all $k \geq 0$.
% %   \begin{equation}
% %   \lambda I \preceq H_k \preceq \Lambda I.
% %   \end{equation}
% \end{assumption}

% \begin{assumption}
%   \label{asm:independent-hessian}
%   $H_k$ depends only on $\{\xi_i\}_{i=1}^{k-1}$.
% \end{assumption}

% \begin{assumption}
%   \label{asm:bounded-step-size}
%   The sequence \(\{\alpha_k > 0\}_{k\geq 0}\) satisfies
%   $
%       \sum_{k=0}^{\infty} \alpha_k = +\infty \, \text{and} \, \sum_{k=0}^{\infty} \alpha_k^2 < +\infty.
%   $
% \end{assumption}

% Note that~\eqref{eq:unbiased-grad-app} is satisfied because we use the mini-batch gradient~\eqref{eq:mini-batch-grad}.
% \Cref{thm:new-L-BFGS-update} shows that \Cref{asm:bounded-hessian} is satisfied.
% \Cref{asm:independent-hessian} is satisfied by~\eqref{eq:BFGS-update_4}.

% Finally, we choose $\alpha_k := c / (k+1)$, where $c > 0$ is a constant, to satisfy \Cref{asm:bounded-step-size}.
Our first result follows from \citet[Theorem~\(2.6\)]{wang2017stochastic}, whose remaining assumptions are satisfied as a consequence of~\eqref{eq:mini-batch-grad}, \Cref{thm:new-L-BFGS-update}, the mechanism of \Cref{alg:VR-SdLBFGS-CHN},~\eqref{eq:BFGS-update_4} and our choice of \(\alpha_k\) below.

\begin{theorem} %[{\protect \citealp[Theorem~\(2.1\)]{wang2017stochastic}}]
  \label{lem:convergence-theorem}
  Assume $m_k = m$ for all $k$, that \Cref{asm:main,asm:stochastic-g-lipschitz,asm:bounded-exp} hold for $\{x_k\}$ generated by \Cref{alg:VR-SdLBFGS-CHN}, and that $\alpha_k := c / (k+1)$ where $0 < c \leq \lambda_{\min} / (L \lambda_{\max})$.
%   If the step size $\alpha_k \leq \frac{\lambda}{L \Lambda} $, for all $k \geq 0$, then
  Then, $\liminf \|\nabla f(x_k)\| = 0$ with probability $1$.
    %   \begin{equation*}
    %       \liminf_{k \to \infty} \|\nabla f(x_k)\| = 0 \quad \text{with probability } 1.
    %   \end{equation*}
  Moreover, there is $M_f > 0$ such that $\E[f(x_k)] \leq M_f$ for all $k$. If we additionally assume that there exists $M_g > 0$ such that $\E_{\xi_k}[\|g(x_k, \xi_k)\|^2] \leq M_g$, then  $\lim \|\nabla f(x_k)\| = 0$ with probability $1$.
    %   \begin{equation*}
    %       \E[f(x_k)] \leq M_f, \quad \forall k \geq 0.
    %   \end{equation*}
\end{theorem}

Our next result follows in the same way from \citet[Theorem~\(2.8\)]{wang2017stochastic}.

\begin{theorem} %[{\protect \citealp[Theorem~\(2.3\)]{wang2017stochastic}}]
  \label{lem:convergence-theorem2}
  Under the assumptions of \Cref{lem:convergence-theorem}, if $\alpha_k = \lambda_{\min} / (L \lambda_{\max}^2) k^{-\beta}$ for all $k > 0$, with $\beta \in (\tfrac{1}{2}, \, 1)$, then,
%   after $N$ iterations,
%     \begin{equation}
%     \frac{1}{T} \sum_{k=1}^{T} \E\left[ \| \nabla f(x_k)\|^2 \right] \leq \frac{2 L (M_f - \kappa_{\text{low}}) \Lambda^2 }{\lambda^2} T^{\beta - 1} + \frac{\sigma^2}{(1 - \beta) m} (T^{-\beta} - T^{-1}).
%     \end{equation}
%     Moreover,
    for any $\epsilon \in (0, \, 1)$, after at most $T = \BigO(\epsilon^{-1 / (1 - \beta)})$ iterations, we achieve
    \begin{equation}
        \frac{1}{T} \sum_{k=1}^{T} \E\left[ \| \nabla f(x_k)\|^2 \right] \leq \epsilon.
    \end{equation}
\end{theorem}

% \begin{itemize}
%   \item lemmas on eigenvalues;
%   \item proof that we satisfy the two assumptions of \citep{cartis-sampaio-toint-2015};
%   \item main complexity result.
% \end{itemize}

% \citep{cartis-sampaio-toint-2015} showed that a direction $d$, whose angle with the steepest descent is controlled by the conditions

% \begin{equation}
% \label{eq:kappa1}
% \nabla f(x)^\top d \leq -\kappa_1 \|\nabla f(x)\|^2,
% \end{equation}
% and
% \begin{equation}
% \label{eq:kappa2}
% \|d\| \leq \kappa_2\|\nabla f(x)\|,
% \end{equation}
% where $\kappa_1$ and $\kappa_2$ are two positive constants, is a suitable descent direction for a Goldstein-Armijo-like line search from $x$.
% We show that a direction $d$ computed using the damped L-BFGS method, satisfies these conditions.

\section{Experimental results}

We compare VARCHEN to SdLBFGS-VR~\citep{wang2017stochastic} and to SVRG~\citep{svrg} for solving a multi-class classification problem.
We train a modified version\footnote{\href{https://colab.research.google.com/github/pytorch/ignite/blob/master/examples/notebooks/Cifar10_Ax_hyperparam_tuning.ipynb}{FastResNet Hyperparameters tuning with Ax on CIFAR10}} of the deep neural network model DavidNet\footnote{\href{https://myrtle.ai/learn/how-to-train-your-resnet-4-architecture/}{https://myrtle.ai/learn/how-to-train-your-resnet-4-architecture/}} proposed by David C. Page, on CIFAR-10 \citep{Krizhevsky09learningmultiple} for $20$ epochs.
Note that we also used VARCHEN and SdLBFGS-VR  to solve a logistic regression problem using the MNIST dataset~\citep{lecun2010mnist} and a nonconvex support-vector machine problem with a sigmoid loss function using the RCV1 dataset~\citep{lewis2004rcv1}.
The performance of both algorithms are on par on those problems because, in contrast with DavidNet on CIFAR-10, they are not highly nonconvex or ill conditioned.

\begin{figure}[t]
    \centering
    \includegraphics[width=.45\linewidth,viewport=16 6 396 269]{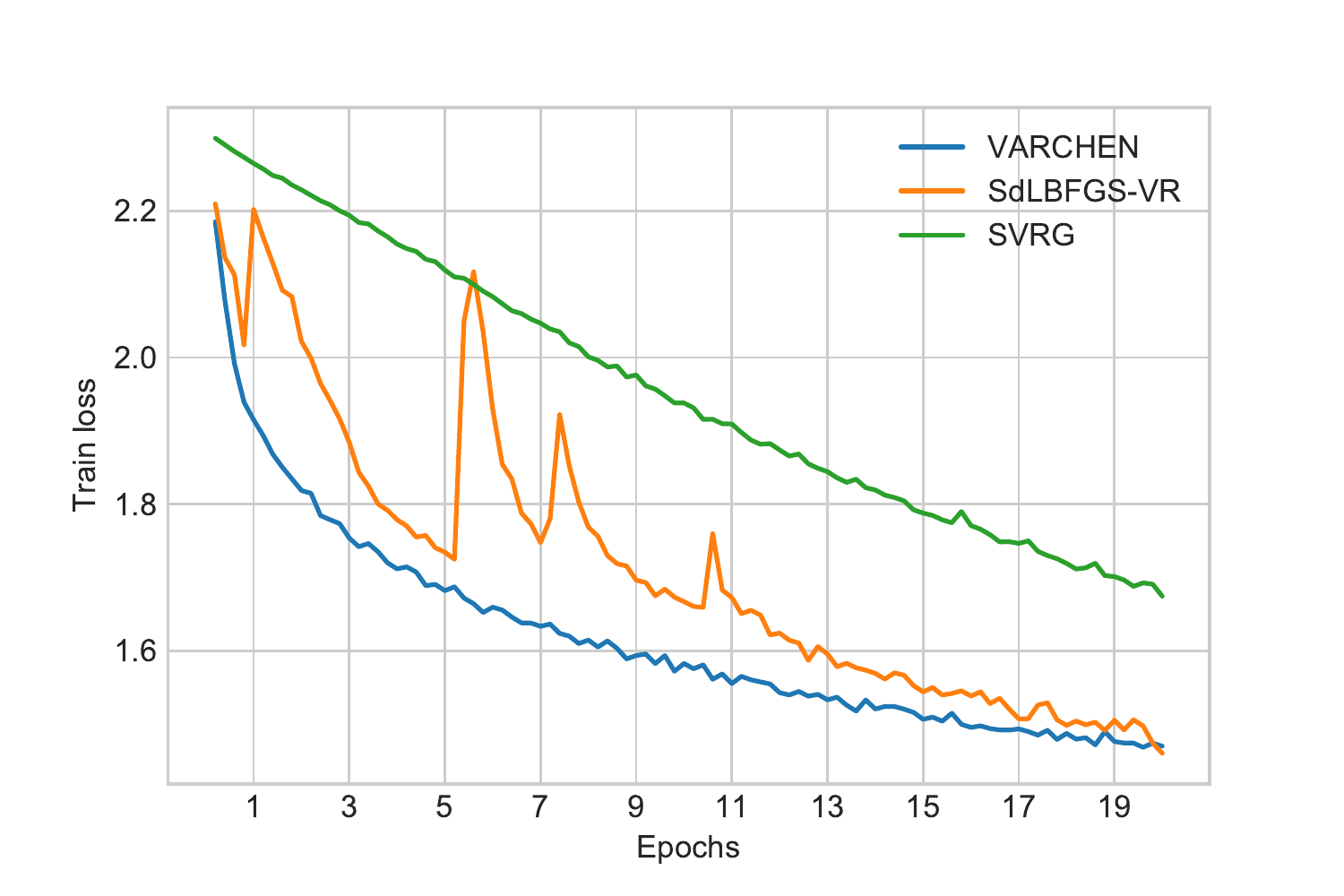}
    \hspace{10mm}
    \includegraphics[width=.45\linewidth,viewport=16 6 396 269]{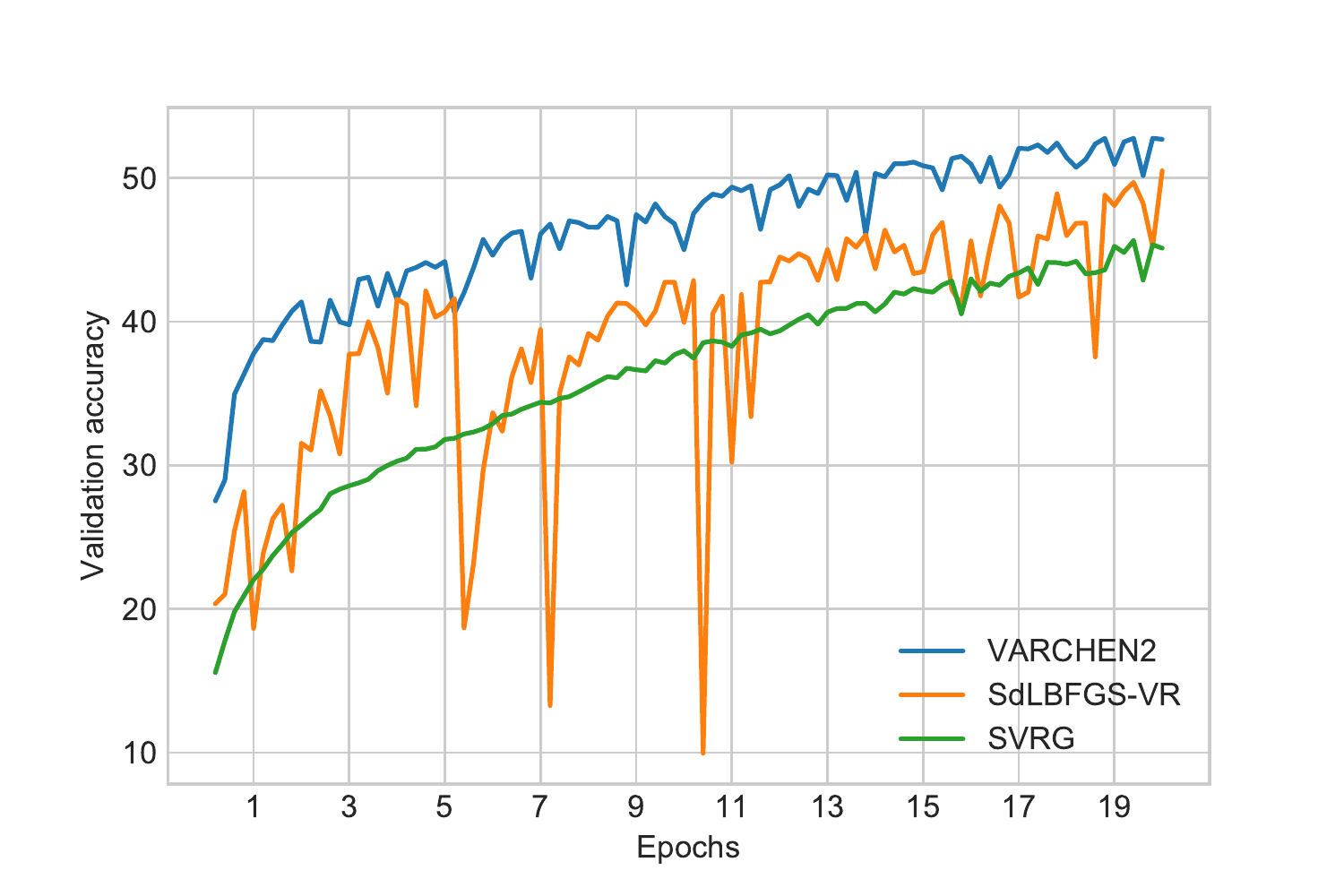}
    % \hfill
    % \includegraphics[width=.325\linewidth,viewport=16 6 394 269]{images/eigenvalues_davnet.pdf}
    \caption{Evolution of the training loss (left) and the validation accuracy (right) for training a modified DavidNet on CIFAR-10. }
    \label{fig:CIFAR-Davnet1}
\end{figure}

\begin{figure}[t]
    \centering
    \includegraphics[width=.45\linewidth,viewport=16 6 396 269]{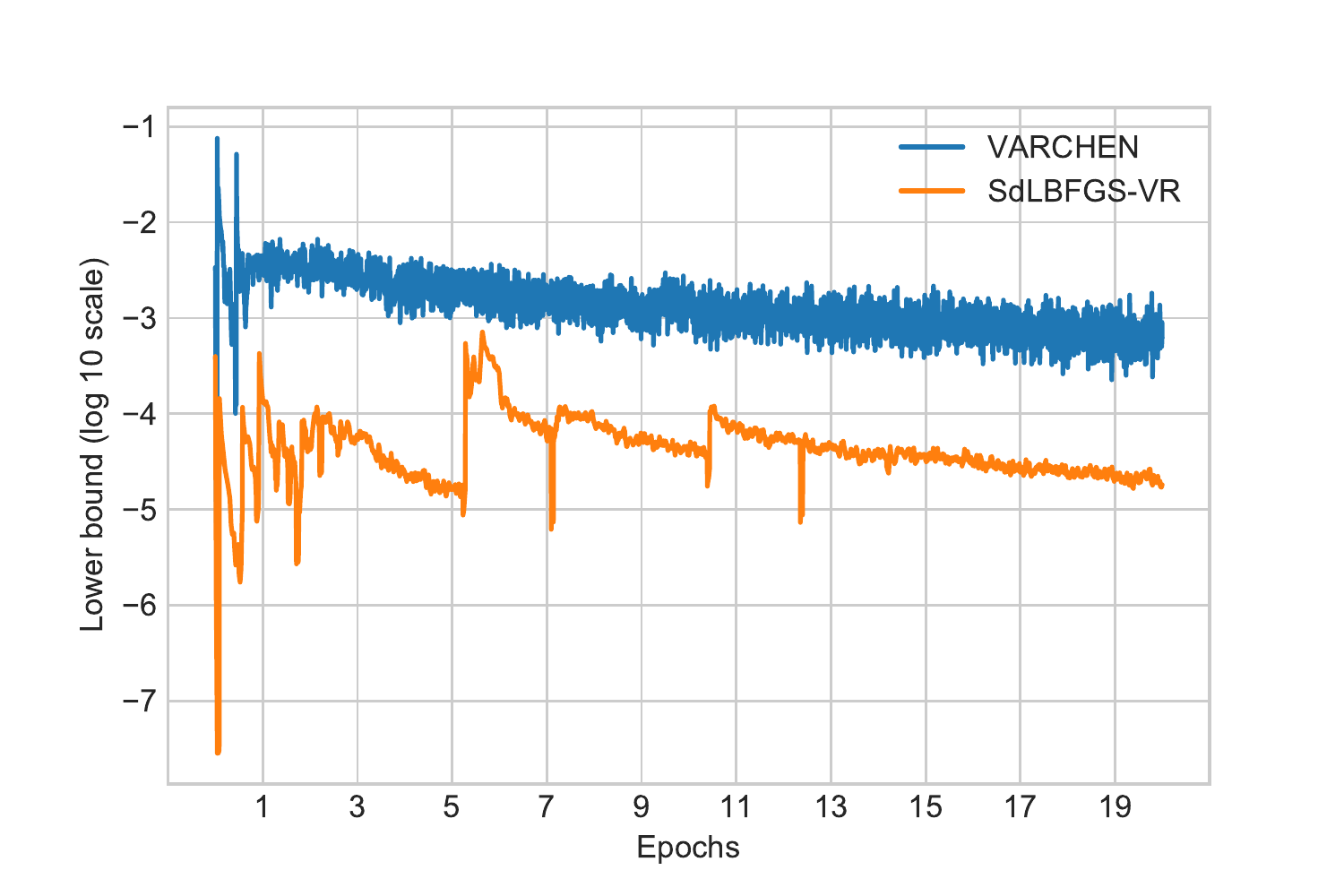}
    \hspace{10mm}
    \includegraphics[width=.45\linewidth,viewport=16 6 396 269]{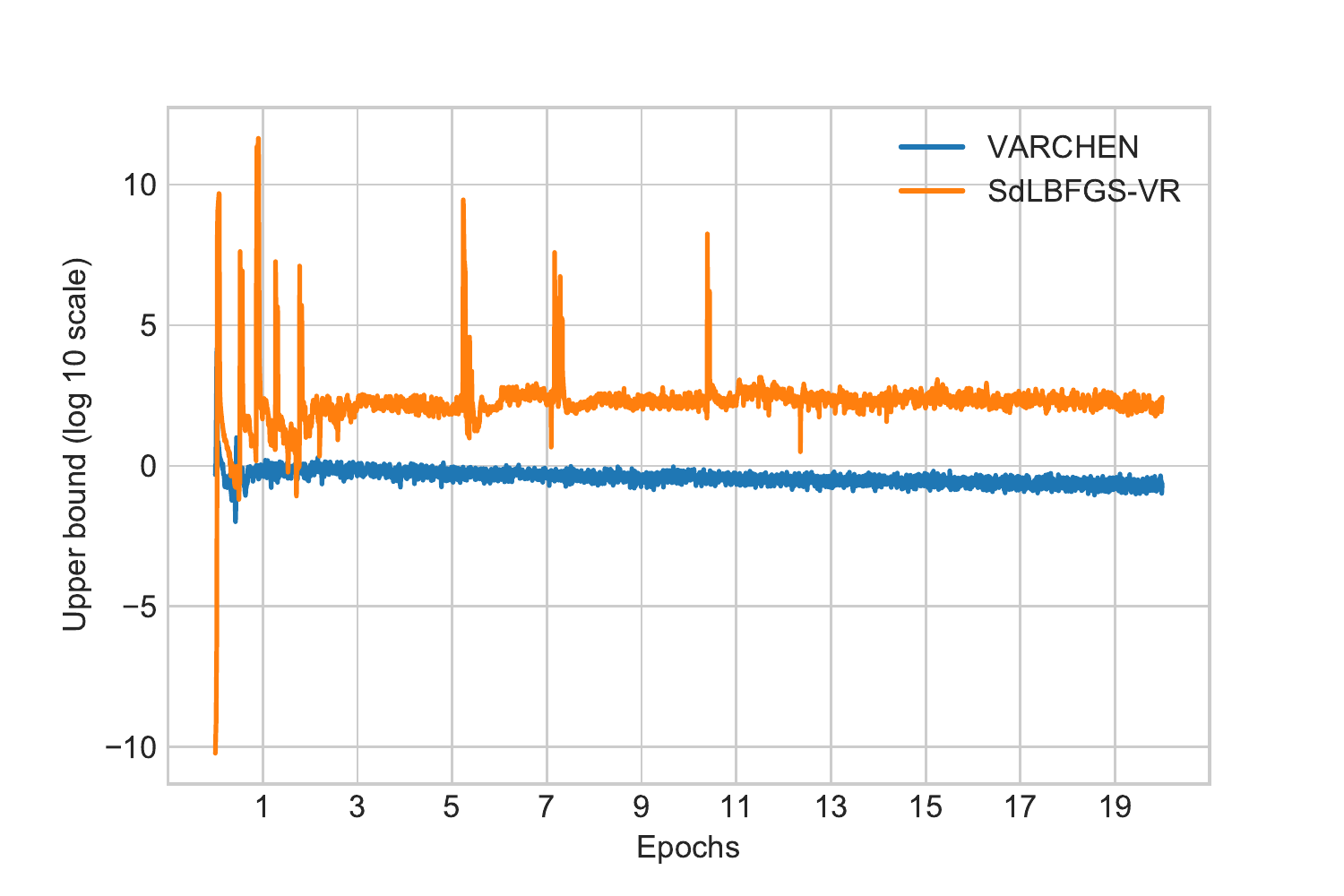}
    % \hfill
    % \includegraphics[width=.325\linewidth,viewport=16 6 394 269]{images/eigenvalues_davnet.pdf}
    \caption{Evolution of the lower bound on the smallest eigenvalue \(\lambda_k\) (left) and the upper bound on the largest eigenvalue \(\Lambda_k\) (right) on a base 10 logarithmic scale for training a modified DavidNet on CIFAR-10. }
    \label{fig:CIFAR-Davnet2}
\end{figure}

\Cref{fig:CIFAR-Davnet1} shows that VARCHEN outperforms SdLBFGS-VR for the training loss minimization task, and both outperform SVRG.
% \todo{need PDF plots instead of PNG so we can better use the bounding box}
VARCHEN has an edge over SdLBFGS-VR in terms of the validation accuracy, and both outperform SVRG.
More importantly, the performance of VARCHEN is more consistent than that of SdLBFGS-VR, displaying a smoother, less oscillatory behaviour.
% \todo{more stable how? less oscillations? should be more precise}
To further investigate this observation, we plot the evolution of \(\Lambda_k\) and \(\lambda_k\) as shown in \Cref{fig:CIFAR-Davnet2}. We see that the estimate of the lower bound on the smallest eigenvalue is smaller for SdLBFGS-VR compared to VARCHEN.
We also notice that the estimate of the upper bound of the largest eigenvalue of $H_k$ takes even more extreme values for SdLBFGS-VR compared to VARCHEN.
The extreme values $\lambda_k$ and $\Lambda_k$ reflect an ill-conditioning problem encountered when using SdLBFGS-VR and we believe that it explains the extreme oscillations in the performance of SdLBFGS-VR.
% \todo{instability = oscillations?}
% Therefore, we demonstrate that our technique, which aims to control the quality of the Hessian approximation, provides a more stable behaviour for highly nonconvex and ill-conditioned problems.

% \begin{figure}[htbp]
%     \centering
%     \includegraphics[scale=0.6]{images/2nd_order_cifar_lower_bound_new.png}
%     \caption{Evolution of the lower bound on the smallest eigenvalues of the inverse Hessian approximation for training a modified DavidNet on CIFAR-10.}
%     \label{fig:CIFAR-Davnet2}
% \end{figure}

% \begin{figure}[htbp]
%     \centering
%     \includegraphics[scale=0.53]{images/2nd_order_cifar_train_loss_new.png}
%     \hspace{5mm}
%     \includegraphics[scale=0.53]{images/2nd_order_cifar_accuracy_new.png}
%     \caption{Evolution of the training loss (left) and accuracy (right) for training a modified DavidNet on CIFAR-10.}
%     \label{fig:CIFAR-Davnet1}
% \end{figure}

% \begin{figure}[htbp]
%     \centering
%     \includegraphics[scale=0.6]{images/2nd_order_cifar_lower_bound_new.png}
%     \caption{Evolution of the lower bound on the smallest eigenvalues of the inverse Hessian approximation for training a modified DavidNet on CIFAR-10.}
%     \label{fig:CIFAR-Davnet2}
% \end{figure}

\section{Conclusion}

We used the stochastic damped L-BFGS algorithm in a nonconvex setting, where there are no guarantees that \(H_k\) remains well-conditioned and numerically nonsingular throughout.
We introduced a new stochastic damped L-BFGS algorithm that monitors the quality of \(H_k\) during the optimization by maintaining bounds on its largest and smallest eigenvalues.
Our work is the first to address the Hessian singularity problem by approximating and leveraging such bounds.
Moreover, we proposed a new initial inverse Hessian approximation that results in  a smoother, less oscillatory training loss and validation accuracy evolution.
% \todo{what evidence do we offer?}
Additionally, we used variance reduction in order to improve the quality of the curvature approximation and accelerate convergence.
Our algorithm converges almost-surely to a stationary point and numerical experiments have shown that it is more robust to ill-conditioned problems and more suitable to the highly nonconvex context of deep learning than SdLBFGS-VR. We consider this work to be a first step towards the use of bounds estimates to control the quality of the Hessian approximation in approximate second-order algorithms. Future work should aim to improve the quality of these bounds and explore another form of variance reduction that consists of adaptive sampling~\citep{jalilzadeh2018variable, bollapragada2018progressive, bollapragada2019adaptive}.
% \todo{Do we talk about the limits here as well?}

% \newpage

\bibliography{bib}
%\newpage
\clearpage
\appendix

% \newpage

\section{Proofs}~\label{app:proofs}
% \firstlemma*

% \begin{lemma}
% %   \label{lem:eigenvalues-update}
%   Let $s$ and $y \in \mathbb{R}^n$ such that $s^\top y \geq \gamma \|s\|^2$ with $\gamma > 0$, and such that $\|y\|\leq L_y \|s\|$, with $L_y > 0$.
%   We denote $\rho$, the inverse of the inner product of these two vectors, i.e. $\rho = \tfrac{1}{s^\top y}$.
%   Let $A =\mu S S^\top  + \rho s s^\top $, where $\mu > 0$, and $S = (I - \rho sy^\top )$.
%   Then,
%   \begin{align}
%     \lambda_{\min}(A) & \geq \min \left( \frac{1}{L_y}, \frac{\mu}{1 + \frac{\mu}{\gamma} L_y^2} \right) > 0,
%     \label{eq:lambda-min-appendix}
%     \intertext{and}
%     \lambda_{\max}(A) & \leq \frac{1}{\gamma} + \max \left( 0, \frac{\mu}{\gamma^2}L_y^2 - \frac{\mu}{1 + \frac{\mu}{\gamma}L_y^2} \right).
%     \label{eq:lambda-max-appendix}
%   \end{align}
% \end{lemma}

\begin{proof}\! \textbf{of \Cref{lem:eigenvalues-update}.} 
First notice that
\begin{equation*}
    \gamma\|s\|^2 \leq s^\top y \leq \|s\|\|y\|,
    \quad \text{and thus} \quad \|s\| \leq \frac{1}{\gamma}\|y\|.
\end{equation*}

Therefore, 

\begin{equation}
    \label{eq:rho-bounded}
    \frac{1}{\|s\|\|y\|} \leq \rho \leq \frac{1}{\gamma}\frac{1}{\|s\|^2}.
\end{equation}

  Since $A$ is a real symmetric matrix, the spectral theorem states that its eigenvalues are real and it can be diagonalized by an orthogonal matrix.
  That means that we can find $n$ orthogonal eigenvectors and $n$ eigenvalues counted with multiplicity.

  Consider first the special case where $s$ and $y$ are collinear, i.e. there exists $\theta > 0$ such that $y = \theta s$.
  Any vector such that $u \in s^\perp$, where $s^\perp = \{x \in \R^n : x^\top s = 0\}$, is an eigenvector of $A$ associated with the eigenvalue $\mu$ of multiplicity $n-1$.
  Moreover, $s^\top y = \theta \|s\|^2=\|s\|\|y\|$, $\rho = 1 / (\theta \|s\|^2)$ and we have
  \begin{equation*}
      As   = \left[\mu \left(I-\rho\theta ss^\top \right)^2 + \rho ss^\top  \right] s
           = \left[\mu \left( 1-\rho\theta\|s\|^2 \right)^2 + \rho\|s\|^2\right] s
           = \rho \|s\|^2 s.
  \end{equation*}
  Let us call $\lambda = \rho\|s\|^2$, the eigenvalue associated with eigenvector $s$.
  From~\eqref{eq:rho-bounded} and  $\|y\|\leq L_y \|s\|$, we deduce that
  \begin{equation*}
      \frac{1}{L_y} \leq \lambda \leq \frac{1}{\gamma}.
  \end{equation*}

  Suppose now that $s$ and $y$ are linearly independent.
  Any $u$ such that $u^\top s = 0 = u^\top y$ satisfies $Au = \mu u$.
  This provides us with a ($n-2$)-dimensional eigenspace $S$, associated to the eigenvalue $\mu$ of multiplicity $n-2$.
  Note that
  \begin{align*}
      As &= \rho \|s\|^2\ (1 + \mu \rho \|y\|^2)s - \rho \|s\|^2 \mu y, \\
      Ay &= s.
  \end{align*}
  Thus neither $s$ nor $y$ is an eigenvector of $A$.
  Now consider an eigenvalue $\lambda$ associated with an eigenvector $u$, such that $u \in S^\perp$.
  Since $s$ and $y$ are linearly-independent, we can search for $u$ of the form $u = s + \beta y$ with $\beta > 0$.
  The condition \(A u = \lambda u\) yields
  \begin{align*}
    \rho \|s\|^2\ (1 + \mu \rho \|y\|^2) + \beta &= \lambda, \\
    -\rho \|s\|^2\mu &= \lambda \beta.
  \end{align*}
  We eliminate $\beta = \lambda - \rho \|s\|^2\ (1 + \mu \rho \|y\|^2)$ and obtain
  \begin{equation*}
    p(\lambda) = 0,
  \end{equation*}
  where
  \begin{equation*}
      p(\lambda) = \lambda^2 - \lambda \rho \|s\|^2\ (1 + \mu \rho \|y\|^2) + \rho \|s\|^2\mu .
  \end{equation*}
  The roots of $p$ must be the two remaining eigenvalues $\lambda_1 \leq \lambda_2$ that we are looking for.
  In order to establish the lower bound, we need a lower bound on $\lambda_1$ whereas to establish the upper bound, we need an upper bound on $\lambda_2$.

  On the one hand, let $l$ be the tangent to the graph of $p$ at $\lambda = 0$, defined by
  \begin{equation*}
      l(\lambda) = p(0) + p'(0)\lambda = \mu \rho \|s\|^2 - \lambda \rho \|s\|^2\left(1 + \mu \rho \|y\|^2\right).
  \end{equation*}
  Its unique root is
  \begin{equation*}
      \bar{\lambda} = \frac{\mu}{1 + \mu \rho \|y\|^2}.
  \end{equation*}
  From~\eqref{eq:rho-bounded} and since $\|y\|\leq L_y \|s\|$, we deduce that
  \begin{equation*}
  \label{eq:lambda-bar-bounded}
    \bar{\lambda} \geq \frac{\mu}{1 + \frac{\mu}{\gamma} \frac{\|y\|^2}{\|s\|^2}} \geq \frac{\mu}{1+ \frac{\mu}{\gamma} L_y^2}.
  \end{equation*}
  Since $p$ is convex, it remains above its tangent, and $\bar{\lambda} \leq \lambda_1$.
  
  Finally, 
  $$\lambda_{\min}(A) \geq \min \left( \frac{1}{L_y}, \frac{\mu}{1+ \frac{\mu}{\gamma} L_y^2} \right) > 0.$$
  
  This establishes the lower bound.

  On the other hand, the discriminant $\Delta = \rho^2 \|s\|^4 (1 + \mu \rho \|y\|^2)^2 - 4 \rho \|s\|^2 \mu$ must be nonnegative since $A$ is real symmetric, and its eigenvalues are real.
  We have
  \begin{equation*}
      \lambda_2 = \frac{ \rho \|s\|^2 (1 + \mu \rho \|y\|^2) + \sqrt{ \rho^2 \|s\|^4 (1 + \mu \rho \|y\|^2)^2 - 4 \rho \|s\|^2 \mu } }{ 2 }.
  \end{equation*}
  For any positive $a$ and $b$ such that $a^2 - b > 0$, we have $\sqrt{a^2-b} \leq a - \tfrac{b}{2a}$. Thus,
  \begin{equation*}
      \lambda_2 \leq \rho \|s\|^2 (1 + \mu\rho\|y\|^2) - \frac{\mu}{1+\mu\rho\|y\|^2}.
  \end{equation*}
  From~\eqref{eq:rho-bounded}, we deduce that
  \begin{equation*}
      \lambda_2 \leq \frac{1}{\gamma} + \frac{\mu}{\gamma^2}\frac{\|y\|^2}{\|s\|^2} - \frac{\mu}{1 + \frac{\mu}{\gamma}\frac{\|y\|^2}{\|s\|^2}}.
  \end{equation*}
  And since $\|y\|\leq L_y \|s\|$, it follows
  \begin{equation*}
      \lambda_2 \leq \frac{1}{\gamma} + \frac{\mu}{\gamma^2}L_y^2 - \frac{\mu}{1 + \frac{\mu}{\gamma}L_y^2}.
  \end{equation*}
  Finally, 
  $$\lambda_{\max}(A) \leq \max
  \left(\frac{1}{\gamma}, \frac{1}{\gamma} + \frac{\mu}{\gamma^2}L_y^2 - \frac{\mu}{1 + \frac{\mu}{\gamma}L_y^2}\right),$$
  which establishes the upper bound.
\end{proof}

% \begin{theorem}
% %   \label{lem:new-L-BFGS-update}
% Let $B_k = H_k^{-1}$ and $H_{k+1}^0$ be two positive definite matrices and $p > 0$.
% If $H_{k+1}$ is obtained by applying $p$ times the damped BFGS update formula with inexact gradient~\eqref{eq:damped-BFGS-update} to $H_{k+1}^0$, then there exists two positive constants $\lambda$ and $\Lambda$ such that
% \begin{equation}
% % \label{eq:eigenvalues-bounds}
%     \lambda_{\min}(H_{k+1}) \geq \lambda \qquad \text{and} \qquad \lambda_{\max}(H_{k+1})\leq \Lambda.
% \end{equation}

% \end{theorem}

\begin{proof}\! \textbf{of \Cref{thm:new-L-BFGS-update}.}
Consider one damped BFGS update using $s$ and $y$ defined in~\eqref{eq:define-y-s} and $\hat{y}_k$ defined in~\eqref{eq:new-ytilde}, i.e, $p = 1$,
% \todo{Use $\hat{V}$ instead of $S$}

\begin{equation*}
H_{k+1} = \hat{V}_k H_{k+1}^0 \hat{V}_k^\top  + \hat{\rho}_k s_k s_k^\top , \quad \text{where} \quad \hat{\rho}_k = 1/s_k^\top  \hat{y}_k, \quad \hat{V}_k = I-\hat{\rho}_k s_k \hat{y}_k^\top.
\end{equation*}

Let \(0 < \mu_1 := \lambda_{\min}(H_{k+1}^0) \leq \mu_2 := \lambda_{\max}(H_{k+1}^0)\).
We have
\begin{equation}
    \label{eq:eigs-update1}
    \lambda_{\min}( \mu_1 \hat{V}_k \hat{V}_k^\top  + \hat{\rho}_ks_ks_k^\top ) \leq
    \lambda_{\min}(H_{k+1}) \leq
    \lambda_{\max}(H_{k+1}) \leq \lambda_{\max}(\mu_2 \hat{V}_k \hat{V}_k^\top  +  \hat{\rho}_ks_ks_k^\top ).
\end{equation}
% \begin{align*}
%     \lambda_{\min}(H_{k+1}) &\geq \lambda_{\min}( \mu_1 \hat{V}_k \hat{V}_k^\top  + \hat{\rho}_ks_ks_k^\top ), \quad \text{where } \mu_1 = \lambda_{\min}(H_{k+1}^0), \\
%     \lambda_{\max}(H_{k+1}) &\leq \lambda_{\max}(\mu_2 \hat{V}_k \hat{V}_k^\top  +  \hat{\rho}_ks_ks_k^\top ), \quad \text{where } \mu_2 = \lambda_{\max}(H_{k+1}^0).
%   \end{align*}
% Because $H_{k+1}^0$ is positive definite, we have $0 < \mu_1 \leq \mu_2$.
Let us show that we can apply \Cref{lem:eigenvalues-update} to 
\begin{equation*}
A_1 := \mu_1 \hat{V}_k \hat{V}_k^\top  + \hat{\rho}_ks_ks_k^\top  \quad \text{and} \quad A_2 := \mu_2 \hat{V}_k \hat{V}_k^\top + \hat{\rho}_ks_ks_k^\top.
\end{equation*}
From~\eqref{eq:new-sTy-bounded}, we obtain
\begin{equation*}
    s_k^\top\hat{y}_k \geq \eta \lambda_{\min}(B_{k+1}^0)\|s_k\|^2 = \frac{\eta}{\lambda_{\max}(H_{k+1}^0)}\|s_k\|^2 = \frac{\eta}{\mu_2}\|s_k\|^2. 
\end{equation*}
\Cref{asm:stochastic-g-lipschitz} yields
\begin{equation*}
    \|\hat{y}_k\| = \|\theta_k y_k + (1-\theta_k) B_{k+1}^0 s_k\|
     \leq \|y_k\|+\|B_{k+1}^0 s_k\|
     \leq (L_g + 1 / \mu_1) \|s_k\|.
\end{equation*}
% then,
% \begin{equation*}
%     \|\hat{y}_k\| \leq  (L_g + \lambda_{\max}((H_{k+1}^0)^{-1}) \|s_k\| = (L_g + \frac{1}{\lambda_{\min}(H_{k+1}^0)}) \|s_k\|.
% \end{equation*}

% We know that for every norm $\|.\|$ on $\R^n$ and every matrix $A \in \R^{n \times n}$, there is a real constant $C_A > 0$, such that $\| A u \|\leq C_A \|u\|$ , for every vector $u \in \R^n$. Let's note this constant $C_{H,k+1}$ for $(H_{k+1}^0)^{-1}$, then we have:

% \begin{equation*}
%     \|\hat{y}_k\| \leq (L_g +   ) \|s_k\|
% \end{equation*}

Therefore, we can first apply \Cref{lem:eigenvalues-update} with $s_k$, $\hat{y}_k$, $\gamma = \eta / \mu_2 > 0$, $L_y = L_g + 1 / \mu_1 > 0$ and $\mu = \mu_1 > 0$ for $A_1$, and apply it again with $\mu = \mu_2 > 0$ for $A_2$.
Let $L_1 := L_g + 1 / \mu_1$.
\Cref{lem:eigenvalues-update} and~\eqref{eq:eigs-update1} yield
  \begin{align*}
    \lambda_{\min}(H_{k+1}) & \geq \min \left( \frac{1}{L_1}, \frac{\mu_1} {1 + \frac{ \mu_1 \mu_2 }{\eta} L_1^2 } \right) > 0, \\
    \lambda_{\max}(H_{k+1}) &\leq \frac{\mu_2}{\eta} + \max \left( 0, \frac{ \mu_2^3 }{ \eta^2} L_1^2 - \frac{ \mu_2 }{ 1 + \frac{\mu_2^2}{\eta} L_1^2 }  \right).
  \end{align*}

Now, consider the case where $p>1$ and let

\begin{equation*}
H_{k+1}^{(h+1)} := \hat{V}_{k-h} H_{k+1}^{(h)} \hat{V}_{k-h}^\top + \hat{\rho}_{k-h} s_{k-h} s_{k-h}^\top,  \quad 0 \leq h \leq p - 1 ,
\end{equation*}

where

\begin{equation*}
H^{(p)}_{k+1} := H_{k+1}, \quad \hat{\rho}_{k-h} = 1 / s_{k-h}^\top \hat{y}_{k-h}, \quad \hat{V}_k = I-\hat{\rho}_{k-h} s_{k-h} \hat{y}_{k-h}^\top.
\end{equation*}

Similarly to the case $p=1$, %let \(0 < \mu_1^{(h)} := \lambda_{\min}(H_{k+1}^{h}) \leq \mu_2^{(h)} := \lambda_{\max}(H_{k+1}^{h})\).
we may write
% \begin{equation}
%     \label{eq:eigs-updatep}
%     \lambda_{\min}( \mu_1^{(h)} \hat{V}_{k-h} \hat{V}_{k-h}^\top + \hat{\rho}_{k-h} s_{k-h} s_{k-h}^\top) \leq
%     \lambda_{\min}(H_{k+1}^{(h+1)}) \leq
%     \lambda_{\max}(H_{k+1}^{(h+1)}) \leq \lambda_{\max}(\mu_2^{(h)} \hat{V}_{k-h} \hat{V}_{k-h}^\top +  \hat{\rho}_{k-h}s_{k-h}s_{k-h}^\top).
% \end{equation}
\begin{alignat*}{2}
    \lambda_{\min}(H_{k+1}^{(h+1)}) &\geq \lambda_{\min}( \mu_1^{(h)} \hat{V}_{k-h} \hat{V}_{k-h}^\top + \hat{\rho}_{k-h} s_{k-h} s_{k-h}^\top), \qquad \mu_1^{(h)} & := \lambda_{\min}(H_{k+1}^{(h)}), \\
    \lambda_{\max}(H_{k+1}^{(h+1)}) &\leq \lambda_{\max}(\mu_2^{(h)} \hat{V}_{k-h} \hat{V}_{k-h}^\top +  \hat{\rho}_{k-h}s_{k-h}s_{k-h}^\top), \qquad   \mu_2^{(h)} & := \lambda_{\max}(H_{k+1}^{(h)}).
  \end{alignat*}
  
Assume by recurrence that $0 < \mu_1^{(h)} \leq \mu_2^{(h)}$.
We show that we can apply \Cref{lem:eigenvalues-update} to 

\begin{equation*}
  A_1^{(h)} := \mu_1^{(h)} \hat{V}_{k-h} \hat{V}_{k-h}^\top + \hat{\rho}_{k-h}s_{k-h}s_{k-h}^\top \quad \text{and} \quad A_2^{(h)} := \mu_2^{(h)} \hat{V}_{k-h} \hat{V}_{k-h}^\top +  \hat{\rho}_{k-h}s_{k-h}s_{k-h}^\top.
\end{equation*}

From~\eqref{eq:new-sTy-bounded}, we have

  \begin{equation*}
    s_{k-h}^\top\hat{y}_{k-h} \geq \eta \lambda_{\min}(B_{k-h+1}^0)\|s_{k-h}\|^2 = \frac{\eta}{\lambda_{\max}(H_{k-h+1}^0)}\|s_{k-h}\|^2.
  \end{equation*}

Using \Cref{asm:stochastic-g-lipschitz},

\begin{equation*}
    \|\hat{y}_{k-h}\| = \| \theta_{k-h} y_{k-h} + (1-\theta_{k-h}) B_{k-h+1}^0 s_{k-h}\| \leq L_g \|s_{k-h}\| + \|B_{k-h+1}^0 s_{k-h}\|,
\end{equation*}

so that

\begin{equation*}
    \|\hat{y}_{k-h}\| \leq (L_g + \frac{1}{\lambda_{\min}(H_{k-h+1}^0)}) \|s_{k-h}\|.
\end{equation*}

We first apply \Cref{lem:eigenvalues-update} with $s = s_{k-h}$, $y = \hat{y}_{k-h}$, $\gamma = \eta / \lambda_{\max}(H_{k-h+1}^0) > 0$, $L_y = L_g + 1 / \lambda_{\min}(H_{k-h+1}^0) > 0$ and $\mu = \mu_1^{(h)} > 0$ for $A_1^{(h)}$, and apply it a second time with $\mu = \mu_2^{(h)} > 0$ for $A_2^{(h)}$. 
Let $L_{k-h+1} := L_g + 1 / \lambda_{\min}(H_{k-h+1}^0)$ and $\gamma_{k-h+1} = \eta / \lambda_{\max}(H_{k-h+1}^0)$.
Then we have

\begin{equation}
\label{eq:lambdaminpr}
\lambda_{\min}(H_{k+1}^{(h+1)}) \geq \min \left( \frac{1}{L_{k-h+1}}, \frac{ \lambda_{\min}(H_{k+1}^{(h)})} {1 + \frac{ \lambda_{\min}(H_{k+1}^{(h)})}{\gamma_{k-h+1}} (L_{k-h+1} )^2 } \right),
\end{equation}

and

\begin{equation}
\label{eq:lambdamaxpr}
\lambda_{\max}(H_{k+1}^{(h+1)}) \leq \frac{1}{\gamma_{k-h+1}} + \max \left( 0, \frac{ \lambda_{\max}(H_{k+1}^{(h)}) }{ \gamma_{k-h+1}^2} L_{k-h+1}^2 - \frac{ \lambda_{\max}(H_{k+1}^{(h)}) }{ 1 + \frac{\lambda_{\max}(H_{k+1}^{(h)})}{\gamma_{k-h+1}} L_{k-h+1}^2 }  \right).
\end{equation}
  
It is clear that we can obtain the lower bound on $\lambda_{\min}(H_{k+1})$ recursively using~\eqref{eq:lambdaminpr}. Obtaining the upper bound on $\lambda_{\max}(H_{k+1})$ using~\eqref{eq:lambdamaxpr} is trickier. However, we notice that inequality~\eqref{eq:lambdamaxpr} implies

% \begin{align}
% \label{eq:lambdamax2pr}
% \lambda_{\max}(H_{k+1}^{(h+1)}) \leq \frac{1}{\gamma_{k-h+1}} +  \frac{ \lambda_{\max}(H_{k+1}^{(h)}) }{ \gamma_{k-h+1}^2} L_{k-h+1}^2 .
% \end{align}

\begin{equation*}
% \label{eq:lambdamaxpr2}
\lambda_{\max}(H_{k+1}^{(h+1)}) \leq \frac{1}{\gamma_{k-h+1}} + \max \left( 0, \frac{ \lambda_{\max}(H_{k+1}^{(h)}) }{ \gamma_{k-h+1}^2} L_{k-h+1}^2 - \frac{ \lambda_{\min}(H_{k+1}^{(h)}) }{ 1 + \frac{\lambda_{\max}(H_{k+1}^{(h)})}{\gamma_{k-h+1}} L_{k-h+1}^2 }  \right).
\end{equation*}

This upper bound is less tight but it allows us to bound $\lambda_{\max}(H_{k+1})$ recursively.
\end{proof}

% \newpage

\section{Algorithms}~\label{app:algorithms}

The Two-loop recursion algorithm for evaluating the Hessian-gradient product is given by Algorithm~\ref{alg:two-loop-recursion}.
% Note that the notations $\lambda_k$ and $\Lambda_k$ do not refer to the $k$-th eigenvalue and that the index $k$ refers to the iteration number only.

\begin{algorithm}[htbp]
  \caption{Two-loop recursion algorithm for Hessian-gradient product computation}
  \label{alg:two-loop-recursion}
  \begin{algorithmic}[1]
    \REQUIRE Current iterate \(x_k\),  $g(x_k,\xi_k)$, $\hat{\rho}_i = 1 / s_i^\top \hat{y}_i$, $s_i$, $\hat{y}_i$ for $i \in \{k-p,\ldots,k-1\}$.
    \STATE Define $g = g(x_k,\xi_k)$
    \FOR{$i = k - 1, k-2, \ldots, k-p$}
      \STATE Compute $\nu_i = \hat{\rho}_i s_i^\top g$
      \STATE Compute $g = g - \nu_i \hat{y}_i$
    \ENDFOR
    \STATE Compute $q = H_k^0 \, g$ using~\eqref{eq:init_HK}.
    \FOR{$i = k-p, k-p+1, \ldots, k-1$}
    \STATE Compute $\mu = \hat{\rho}_i \hat{y}_i^\top q$
    \STATE Compute $ q = q + (\nu_i - \mu) s_i $
    \ENDFOR
    \STATE Return $q = H_k \, g(x_k,\xi_k)$
  \end{algorithmic}
\end{algorithm}

\section{Experimental Setting}

CIFAR-10 is a dataset that contains $60,000$ colour images with labels in $10$ classes (airplane, automobile, bird, cat, deer, dog, frog, horse, ship, and truck), each containing $6,000$ images. We use $50,000$ images for the training task and $10,000$ images for the validation task.

Details of the experiments:
\begin{itemize}
    \item We apply our Hessian norm control using the bound on the maximum and the minimum eigenvalues of $H_{k+1}$, where the latter is equivalent to controlling $\|B_{k+1}\|$;
    \item In the definition of $H_{k+1}^0$ in~\eqref{eq:init_HK}, we choose $\underline{\gamma}_{k+1}$ and $\overline{\gamma}_{k+1}$ constant;
    \item The numerical values for all algorithms are the ones that yielded the best results among all sets of values that we experimented with. 
\end{itemize}

Numerical values:
\begin{itemize}
    \item For all algorithms: we train the network for $20$ epochs and use a batch size of $256$ samples;
    \item For SVRG, we choose a step size equal to $0.001$;
    \item For both SdLBFGS-VR and \Cref{alg:VR-SdLBFGS-CHN}, the memory parameter $p = 10$, the minimal scaling parameter $\underline{\gamma}_{k+1} = 0.1$ for all $k$, the constant step size $\alpha_k = 0.1$ and $\eta = 0.25$;
    \item For \Cref{alg:VR-SdLBFGS-CHN}, we use a maximal scaling parameter $\overline{\gamma}_{k+1} = 10^5$ for all $k$, a lower bound limit $\lambda_{\min} = 10^{-5}$ and an upper bound limit $\lambda_{\max} = 10^{5}$.
\end{itemize}

\end{document}